\documentclass[conference]{IEEEtran}

\ifCLASSINFOpdf
\else
\fi
\usepackage{cite}
\usepackage{amsmath,amssymb,amsfonts}
\usepackage{comment}

\usepackage[ruled,vlined]{algorithm2e}

\SetKwInput{Input}{Input}
\SetKwInput{Output}{Output}

\usepackage{pifont}
\newcommand{\cmark}{\ding{51}}
\newcommand{\xmark}{\ding{55}}

\usepackage{tabularx}
\usepackage{array}      
\usepackage{booktabs}   

\usepackage{multirow}
\usepackage{graphicx}
\usepackage{textcomp}
\usepackage{xcolor}
\def\BibTeX{{\rm B\kern-.05em{\sc i\kern-.025em b}\kern-.08em
    T\kern-.1667em\lower.7ex\hbox{E}\kern-.125emX}}

\SetAlCapSkip{0.2em}      
\SetAlCapHSkip{0pt}
\SetAlgoSkip{0.1em}       
\SetAlgoNlRelativeSize{-2}
\begin{document}

%
\title{ImageHD: Energy-Efficient On-Device Continual Learning of Visual Representations via Hyperdimensional Computing}

\author{
\IEEEauthorblockN{Jebacyril Arockiaraj, Dhruv Parikh, Viktor Prasanna}  
\IEEEauthorblockA{University of Southern California, Los Angeles, CA, USA \\  
Email: \{arockiar, dhruvash, prasanna\}@usc.edu}  
}

\maketitle

\begin{abstract}
On-device continual learning (CL) enables edge devices to adapt to non-stationary data streams without offline retraining, which is critical for real-time edge AI. However, most existing CL methods rely on backpropagation-based updates or exemplar-heavy classifiers, incurring high compute, memory, and latency overheads that hinder deployment on resource-constrained devices. Hyperdimensional Computing (HDC) offers an alternative by enabling fast, non-iterative online updates. When paired with a lightweight convolutional neural network (CNN) feature extractor, HDC supports efficient on-device adaptation with strong visual representations. Despite this progress, prior HDC-based continual learning systems typically employ multi-tier memory hierarchies and complex cluster management, which complicate deployment on resource-constrained hardware platforms.

In this paper, we propose ImageHD, an FPGA accelerator for on-device continual learning of visual data based on HDC, designed to address these limitations. ImageHD targets streaming continual learning under strict latency and on-chip memory constraints, without expensive iterative optimization. At the algorithmic level, we introduce a hardware-optimized continual learning method that bounds total class exemplars through a unified exemplar memory and a hardware-efficient cluster merging strategy, while integrating a quantized CNN front-end to reduce edge deployment overhead without sacrificing accuracy. At the system level, ImageHD is realized as a streaming dataflow architecture on the AMD Zynq ZCU104 FPGA, integrating hyperdimensional encoding, similarity search, and bounded cluster management using word-packed binary hypervector representations to enable massively parallel bitwise operations within tight on-chip resource budgets. Experimental results on the CORe50 dataset demonstrate up to $40.4\times$ ($4.84\times$) speedups and $383\times$ ($105.1\times$) energy efficiency gains over optimized CPU (GPU) baselines, establishing HDC-enabled continual learning as a practical foundation for real-time, on-device lifelong learning systems.
\end{abstract}

\begin{IEEEkeywords}
Continual learning, hyperdimensional computing, FPGA acceleration, on-device learning
\end{IEEEkeywords}


%
\IEEEpeerreviewmaketitle

\section{Introduction}
Continual Learning (CL) ~\cite{Chen2018Lifelong, Aljundi2017ExpertGate, Chaudhry2018AGEM, Parisi2019CLReview, Silver2002TaskRehearsal}, is the ability of an artificial intelligence (AI) system to incrementally acquire knowledge from sequence of data streams, similar to biological learning systems, without training from scratch~\cite{Hadsell2020Continual}. This capability is important for edge applications, where devices continuously encounter new visual data and must adapt on-device without reliance on cloud retraining. Scenarios include autonomous drones, robotic inspection systems, IoT sensors, etc. where network connectivity may be limited, and energy and latency budgets are constrained.


Table~\ref{tab:cl_comprehensive} summarizes continual learning methods and highlights key trade-offs relevant to edge deployment. Approaches such as~\cite{Hayes2020DeepSLDA} and~\cite{AILIS} rely on supervised learning and require labeled data. However, in realistic continual learning scenarios, manual annotation is often impractical, particularly when new data is generated on-the-fly~\cite{Fini2022CaSSLe}. Self-supervised methods, including ~\cite{Fini2022CaSSLe} and ~\cite{Madaan2022LUMP}, eliminate the need for labels but depend on backpropagation-based training, resulting in high computational cost and making them unsuitable for resource-constrained edge devices. These limitations motivate learning paradigms that avoid both labeled supervision and gradient-based optimization while maintaining the capability for continual learning.


\newcolumntype{L}{>{\raggedright\arraybackslash}X}
\newcolumntype{C}{>{\centering\arraybackslash}X}

\newcolumntype{L}{>{\raggedright\arraybackslash}X}
\newcolumntype{C}{>{\centering\arraybackslash}X}

\begin{table}[t]
\centering
\caption{Comparison of representative continual (visual) learning pipelines from the perspective of \emph{on-device deployability}.}
\label{tab:cl_comprehensive}
\scriptsize
\setlength{\tabcolsep}{3.5pt}
\renewcommand{\arraystretch}{1.1}
\begin{tabularx}{\columnwidth}{L L L L L L}
\toprule
\textbf{Method} &
\textbf{Sup.} &
\textbf{Backbone} &
\textbf{Update Mechanism} &
\textbf{Bounded} &
\textbf{HW} \\
\midrule

Deep SLDA~\cite{Hayes2020DeepSLDA} &
Sup. &
ResNet-18 &
Class statistics (SLDA) &
\xmark &
CPU/GPU \\

AILIS~\cite{AILIS} &
Sup. &
VGG-16 &
Exemplar-based KNN &
\xmark &
FPGA \\

CaSSLe~\cite{Fini2022CaSSLe} &
SSL &
ResNet-18/50 &
Backprop + distill &
\xmark &
GPU \\

LUMP~\cite{Madaan2022LUMP} &
SSL &
ResNet-18 &
Backprop + replay &
\xmark &
GPU \\

LifeHD~\cite{LifeHD} &
Unsup. &
MobileNetV2 &
HDC + spectral merge &
\cmark\textsuperscript{$\dagger$} &
CPU/GPU \\

\textbf{ImageHD (Ours)} &
Unsup. &
INT8 MobileNetV2 &
HDC + kMeans++ merge &
\textbf{\cmark} &
FPGA \\

\bottomrule
\end{tabularx}

{\footnotesize \textsuperscript{$\dagger$}\cite{LifeHD} bounds memory via multi-tier storage and spectral merging, introducing additional control state and cubic-time consolidation.}
\end{table}

Hyperdimensional Computing (HDC) is a brain-inspired computing paradigm that represents information using low-precision vectors, often comprising thousands of dimensions~\cite{Kanerva2009HDC}. Unlike deep neural networks, HDC does not rely on complex iterative training and often generalizes well with minimal data and computation. Even though~\cite{DistHD} prior work introduces a dynamic encoding technique to improve accuracy, HDC accuracy on complex visual tasks still lags behind that of deep neural networks~\cite{10.1145/3665314.3670852, HDCvsNN}. 

~\cite{LifeHD} represents a pioneering effort that bridges this gap by combining a frozen convolutional neural network feature extractor with HDC-based encoding to enable continual learning on visual workloads. To mitigate catastrophic forgetting, where new data overwrites previously learned representations, ~\cite{LifeHD} maintains short-term memory (STM) and long-term memory (LTM), where class hypervectors accumulated in STM are periodically merged into LTM using spectral graph clustering. While effective, this design introduces several limitations. First, the spectral graph clustering introduces a cubic-time complexity ($O(N^3)$) due to the eigen decomposition step. Second, separate short-term and long-term memory structures increase memory footprint and control complexity. Finally, existing implementations utilize general-purpose computing platforms, such as GPUs, which have been shown to be inefficient for HDC workloads due to their high dynamic power consumption, limited memory bandwidth utilization, and lack of hardware specialization~\cite{HyperGRAF}.

To address these limitations, we present ImageHD, a domain-specific FPGA accelerator co-designed for efficient on-device continual learning. Our main contributions include:

\begin{itemize}
    \item Clustering: We exploit the quasi-orthogonality of high-dimensional representations to replace cubic-complexity spectral clustering with a hardware-optimized kMeans++ algorithm. This reduces the algorithmic complexity of memory updates from $O(N^3)$ to linear time ($O(N)$), which results in a speedup of 8.6$\times$.
    \item Unified Exemplar Memory: We replace multi-tier short-term and long-term memory structures with a single, consolidated cluster memory that is explicitly bounded to fit on-chip resources, simplifying control logic and improving hardware efficiency.
    \item Quantization: We leverage the inherent robustness of hyperdimensional representations to deploy an INT8-quantized CNN feature extractor. This design choice significantly reduces the memory footprint of the neural feature extractor, without compromising continual learning accuracy.
    \item Streaming Architecture: We design a streaming dataflow FPGA architecture on the AMD Zynq ZCU104 that combines a layer-wise pipelined, tile-based quantized CNN front-end with a streaming hyperdimensional encoding and similarity engine operating on word-packed binary hypervectors, and a hardware-efficient cluster merge unit that bounds on-chip exemplar growth.
    \item System Efficiency: We evaluate ImageHD on the Core-50 dataset, demonstrating $40.4\times$ and $4.84\times$ speedups and $383\times$ and $105.1\times$ energy efficiency gains over CPU and GPU baselines, respectively, validating its suitability for real-time edge adaptation.
\end{itemize}
\section{On-Device Continual Learning: \\ Motivation and Challenges}
\label{sec: odcl}

\subsection{Definition}
\label{subsec: odcl-def}
On-device AI models refer to AI systems that are trained and deployed directly on edge devices. These models perform data processing and inference locally without the need to transmit data to the cloud~\cite{10.1145/3450494}. On-Device Continual Learning (CL) extends this by requiring the model to not only run locally but to continuously update its knowledge using streaming data~\cite{hayes2022online}. This allows the device to learn new tasks or adapt to changing environments in real-time. In the context of On-Device CL for images, the system processes a continuous input stream $\mathcal{S} = \{(\mathbf{x}_t, y_t)\}_{t=1}^\infty$, where $\mathbf{x}_t \in \mathbb{R}^{H \times W \times C}$ represents raw pixel data. The objective is to map $\mathbf{x}_t$ to a dynamic label set $\mathcal{Y}_t$ while strictly adhering to hardware-imposed constraints~\cite{hayes2022online}.

\subsection{Motivation}
\label{subsec: odcl-motiv}

The growing volume of data motivates learning at the edge, as transmitting large volumes of raw data to the cloud incurs substantial latency and bandwidth overheads~\cite{edge_intelligence}. Furthermore, centralized processing raises privacy concerns in sensitive applications such as healthcare and surveillance, motivating On-Device CL that processes information directly at source~\cite{article_Liu}. Beyond systems-level considerations, continual learning is increasingly regarded as a core capability for modern machine learning systems operating in open-world settings, where data distributions shift and new concepts emerge over time without explicit task boundaries~\cite{mai2022online,graldi2025importance,le2024mixture,ackermann2024transferability,task-free-cl,qi2024interactive,liu2025c,liu2025continual,Yu_2025_CVPR}.


\subsection{Challenges}
\label{subsec: odcl-challenge}
\textbf{Catastrophic Forgetting:} Sequential training leads to catastrophic forgetting, where new data overwrites previously learned representations. Mitigating this typically requires Replay, where the system stores and repeatedly accesses past data~\cite{LopezPaz2017GEM}. However, maintaining such a buffer for high-dimensional data (like images) requires significant memory capacity, which is typically unavailable in edge devices.

\textbf{Latency:} On-Device CL for vision tasks is critical for robots, vehicles, and smart appliances, where the system must continuously adapt to changing inputs. However, if the computational time for learning exceeds the data arrival rate, the system is forced to discard frames~\cite{hayes2022online}, resulting in data loss that directly degrades both learning accuracy and real-time responsiveness. This timeliness requirement is also recognized in prior work on online continual learning for edge devices~\cite{10.1145/3696410.3714796}.

\textbf{Hardware:} Balancing these conflicting demands presents a significant challenge for standard hardware. CPUs often lack the parallelism required for real-time learning, while GPUs exceed the power budget of embedded systems due to high static power. In contrast, \textbf{FPGAs} offer a promising solution by leveraging the architectural flexibility and parallelism required to execute complex learning algorithms efficiently within the strict energy budget of edge devices~\cite{fpga_adv}.


\section{Preliminaries}

\subsection{Hyperdimensional Computing}
\label{subsec:hdc_prelim}
Hyperdimensional Computing (HDC) represents information using high-dimensional binary vectors, referred to as hypervectors (HVs). In this work, all HVs are defined in the binary space $\{0,1\}^{D}$, where $D$ denotes the dimensionality. HDC constructs composite representations using a small set of algebraic operations. \textbf{Binding} associates two HVs using an element-wise XOR operation, denoted by $\odot$, i.e., $\mathbf{h}_{\text{bind}} = \mathbf{a} \odot \mathbf{b}$. This operation produces a hypervector dissimilar to both operands while preserving invertibility. \textbf{Bundling} aggregates a set of HVs $\{\mathbf{h}_i\}$ using an operator denoted by $\oplus$, written as $\bigoplus_i \mathbf{h}_i$. Bundling corresponds to element-wise accumulation followed by thresholding, which performs a majority vote in each dimension to produce a binary HV. \textbf{Permutation} encodes positional or structural information by applying a fixed bijective reordering of dimensions, denoted as $\mathbf{h}_{\text{perm}} = \pi^{(\rho)}(\mathbf{h})$, where $\rho$ specifies the permutation or cyclic shift applied to the HV. Similarity between two binary hypervectors $\mathbf{h}_1, \mathbf{h}_2 \in \{0,1\}^{D}$ is measured using their normalized Hamming distance. 

\section{Computationally Efficient Continual Learning in ImageHD}\label{section-4}


\subsection{Hybrid CNN-HDC Encoding}
\label{subsec:imagehd_encoding}

ImageHD adopts a hybrid CNN-HDC continual learning pipeline inspired by prior work~\cite{LifeHD, AILIS, Hayes2020DeepSLDA, Fini2022CaSSLe, Madaan2022LUMP}. Given an input image $\mathbf{x}$, a lightweight pre-trained (quantized) convolutional neural network (CNN) feature extractor $f_{\text{feat}}(\cdot)$ produces a compact feature vector $\mathbf{z} = f_{\text{feat}}(\mathbf{x}) \in \mathbb{R}^{F}$. 
The CNN front-end is responsible for extracting semantically meaningful visual representations, while all continual adaptation is performed downstream in the HDC space. The feature vector $\mathbf{z}$ is encoded into a binary hypervector $\mathbf{h} \in \{0,1\}^{D}$ using an ID-level encoding scheme \cite{Kanerva2009HDC} with two fixed random lookup memories: a \emph{level table} $\mathcal{L} \in \{0,1\}^{L \times D}$ and a \emph{position table} $\mathcal{P} \in \{0,1\}^{F \times D}$. Each scalar feature $z_i$ is quantized to a discrete level index $\ell_i = \mathrm{Quantize}(z_i) \in \{1,\ldots,L\}$, which selects a corresponding level hypervector $\mathcal{L}[\ell_i]$. The positional identity of feature $i$ is captured by the position hypervector $\mathcal{P}[i]$. These two components are bound to form a feature hypervector $\mathbf{u}_i = \mathcal{P}[i] \odot \mathcal{L}[\ell_i]$, and all feature hypervectors are aggregated via bundling, $\mathbf{h} = \bigoplus_{i=1}^{F} \mathbf{u}_i$. The resulting sample hypervector $\mathbf{h}$ serves as the unified representation for downstream similarity search, admission decision (novelty detection), and cluster updates in the continual learning process.

\subsection{Quantized Neural Front-End}
\label{subsec:quantization}

A key characteristic of HDC is its robustness to noise. Prior work~\cite{LaplaceHD} has shown that HDC representations remain effective even when a substantial fraction of hypervector bits are corrupted, due to high dimensional encoding. Motivated by this, we posit that quantization noise introduced in the neural feature extractor can be tolerated by the downstream HDC encoder without CL performance degradation. Accordingly, we apply post-training static quantization to the pretrained MobileNetV2 feature extractor. 
Weight quantization parameters are derived from the pretrained model, while activation quantization is calibrated using warm-up batches from the continual learning stream, which also initialize class HVs. 


\subsection{Memory and Computationally Efficient On-Device Learning Algorithm}
\label{subsec:end_to_end_algorithm}

\begin{algorithm}[t]
\caption{End-to-End Continual Learning in ImageHD}
\label{alg:imagehd_end2end}
\footnotesize
\SetAlgoLined
\DontPrintSemicolon

\KwIn{
Stream $\{\mathbf{x}_t\}_{t=1}^{T}$; feature extractor $f_{\text{feat}}$; \\
level table $\mathcal{L} \in \{0,1\}^{L\times D}$; position table $\mathcal{P} \in \{0,1\}^{F\times D}$; \\
novelty parameter $\beta$; update rate $\alpha$; merge period $T_{\text{merge}}$; capacity $C_{\max}$; merge start $T_0$.
}
\KwOut{
Predicted cluster label $\hat{y}_t$ and updated cluster set $\mathcal{C}$.
}

\textbf{State:} clusters $\mathcal{C}=\{c\}$, where each $c$ stores $(\mathbf{h}_c,\mu_c,\hat{\sigma}_c)$ \;

\For{$t \gets 1$ \KwTo $T$}{
    \tcp{S1: Feature extraction}
    $\mathbf{z}_t \gets f_{\text{feat}}(\mathbf{x}_t)$ \;

    \tcp{S2: HDC encoding}
    $\boldsymbol{\ell}_t \gets \mathrm{Quantize}(\mathbf{z}_t)$\;
    $\mathbf{q}_t \gets \bigoplus_{i=1}^{F}\left(\mathcal{P}[i]\odot \mathcal{L}[\ell_{t,i}]\right)$\;

    \tcp{S3: Similarity search}
    $s_c \gets \mathrm{sim}(\mathbf{q}_t,\mathbf{h}_c)$ \;
    $(s^\star, c^\star) \gets \arg\max_{c} s_c$ \;

    \tcp{S4: Admission decision}
    $\theta_{c^\star} \gets \mu_{c^\star} - \beta\,\hat{\sigma}_{c^\star}$ \;
    \eIf{$s^\star < \theta_{c^\star}$}{
        create $c_{\text{new}}$ with $\mathbf{h}_{c_{\text{new}}}\gets \mathbf{q}_t$ \;
        $\mathcal{C} \gets \mathcal{C} \cup \{c_{\text{new}}\}$ \;
    }{
        $\mathbf{h}_{c^\star} \gets \mathbf{h}_{c^\star} \oplus \mathbf{q}_t$ \;
    }

    \tcp{S5: Cluster merge}
    \If{$t \ge T_0$ \textbf{and} $t \bmod T_{\text{merge}}=0$ \textbf{and} $|\mathcal{C}|>C_{\max}$}{
        $\mathcal{C} \gets \textsc{MergeClusters}(\mathcal{C},C_{\max})$ \;
    }
}
\end{algorithm}

\begin{algorithm}[t]
\caption{\textsc{MergeClusters}: Top-$M$ kMeans++ with HV Refinement}
\label{alg:imagehd_mergeclusters}
\footnotesize
\SetAlgoLined
\DontPrintSemicolon

\Input{$\mathcal{C}=\{\mathbf{h}_1,\ldots,\mathbf{h}_K\}$; target $K'$; Top-$M$ size $M$; refinement iterations $I$}
\Output{$\mathcal{C}'=\{\mathbf{c}_1,\ldots,\mathbf{c}_{K'}\}$}

$u \gets \textsc{UniformRand}(1,\ldots,K)$; $\mathbf{c}_1 \gets \mathbf{h}_u$ \;
$\mathbf{d}[i] \gets \mathrm{Ham}(\mathbf{h}_i,\mathbf{c}_1)\;\forall i$ \;

\For{$m \gets 2$ \KwTo $K'$}{
    $\mathcal{T}_m \gets \textsc{SelectTopM}(\mathbf{d},M)$ \;
    $u \gets \textsc{UniformSample}(\mathcal{T}_m)$ \;
    $\mathbf{c}_m \gets \mathbf{h}_u$ \;
    $\mathbf{d}[i] \gets \min(\mathbf{d}[i],\mathrm{Ham}(\mathbf{h}_i,\mathbf{c}_m))\;\forall i$ \;
}
$\{\mathbf{c}_m\}_{m=1}^{K'} \gets \textsc{HVRefine}(\mathcal{C},\{\mathbf{c}_m\},I)$ \;
\Return{$\mathcal{C}'$} \;
\end{algorithm}


Algorithm~\ref{alg:imagehd_end2end} implements single-pass streaming continual learning with a bounded cluster set. Each sample runs feature extraction (S1), HDC encoding (S2) to form a binary query HV by binding quantized features with position HVs and bundling, then performs similarity search over all cluster HVs using the metric in Section~\ref{subsec:hdc_prelim} (S3). An admission decision (S4) either updates the matched cluster (bundling + $(\mu,\hat{\sigma})$ running-statistics with rate $\alpha$) or creates a new cluster (for novel samples).

A key distinction of ImageHD relative to prior work~\cite{LifeHD} is its \emph{cluster management strategy}. The existing HDC-based continual learning pipelines rely on multi-tier exemplar memories and costly global consolidation (e.g., spectral or graph-based clustering). Spectral clustering is typically required for non-convex structures that are not well captured by simple distance metrics. However, in HDC, the hypervectors exhibit quasi-orthogonality~\cite{zhou2025factorhdhyperdimensionalcomputingmodel}. As a result, unrelated hypervectors have low cosine similarity. This makes the cluster structure in HDC embeddings well-suited to simpler distance-based clustering.

ImageHD maintains a \emph{single-tier, bounded cluster memory} throughout execution. This eliminates additional control state and irregular memory accesses, which are critical impediments to efficient on-chip deployment. When the cluster count exceeds capacity, ImageHD invokes a hardware-friendly consolidation step, \textsc{MergeClusters} (S5), which performs distance-based merging directly in HV space, exploiting the approximate linear separability of high-dimensional binary hypervectors without eigen-decompositions.

\textsc{MergeClusters} (Algorithm~\ref{alg:imagehd_mergeclusters}) consolidates a cluster set $\mathcal{C}$ into $K'$ representatives via a two-stage procedure. Centroid initialization uses a simplified kMeans++ strategy: the first centroid is selected uniformly at random, and subsequent centroids are sampled uniformly from the Top-$M$ farthest points based on minimum Hamming distance to the current centroid set. 
A small, fixed number of HV-space refinement steps then follow, where each HV is assigned to its nearest centroid and centroids are updated via HDC-style bundling with majority thresholding.

\section{Problem Definition}
\label{subsec:fpga_problem_definition}

The objective of ImageHD is to enable \emph{fully on-device continual learning} for streaming image inputs on resource-constrained edge FPGAs. The system executes the complete learning pipeline (Algorithm~\ref{alg:imagehd_end2end}) in a single pass, incrementally assigning samples to existing clusters or instantiating new ones without retraining or data replay. 
Deployment is constrained by limited on-chip memory, which must store HDC tables and a bounded set of cluster hypervectors, while neural weights are streamed from external memory. 
\section{Accelerator Design}

\begin{figure}[ht]
    \centering
    \includegraphics[width=0.85\columnwidth]{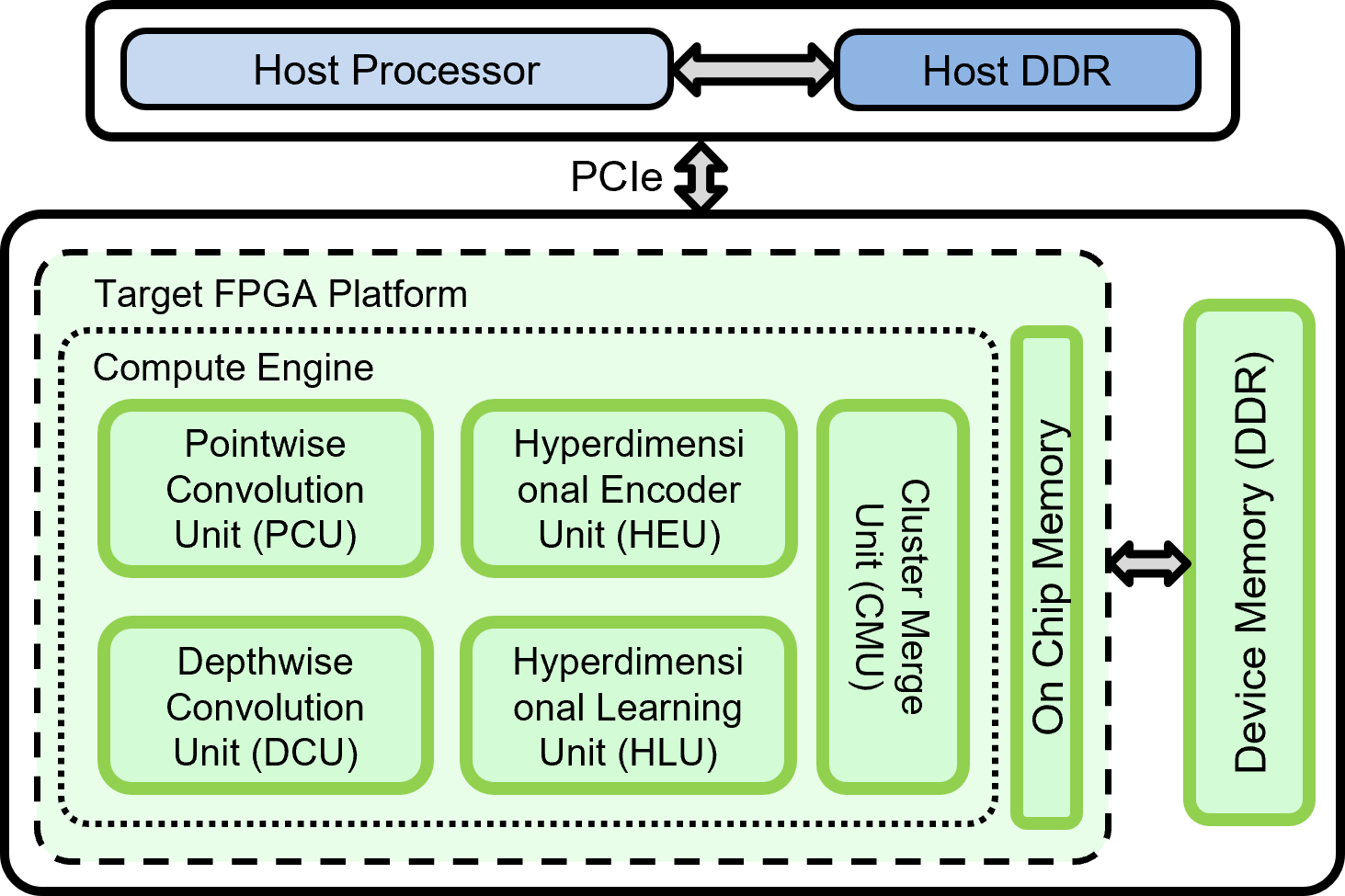}
    \caption{Overview of ImageHD accelerator.}
    \label{fig:overview}
\end{figure}

Figure~\ref{fig:overview} illustrates the overall architecture of ImageHD, an FPGA accelerator for on-device continual visual learning. ImageHD comprises five specialized compute units: (i) Pointwise Convolution Unit (PCU); (ii) Depthwise Convolution Unit (DCU); (iii) Hyperdimensional Encoding Unit (HEU); (iv) Hyperdimensional Classifier Unit (HCU); and (v) Cluster Merge Unit (CMU). We cover each unit in detail below.

\begin{figure}
    \centering
    \includegraphics[width=\columnwidth]{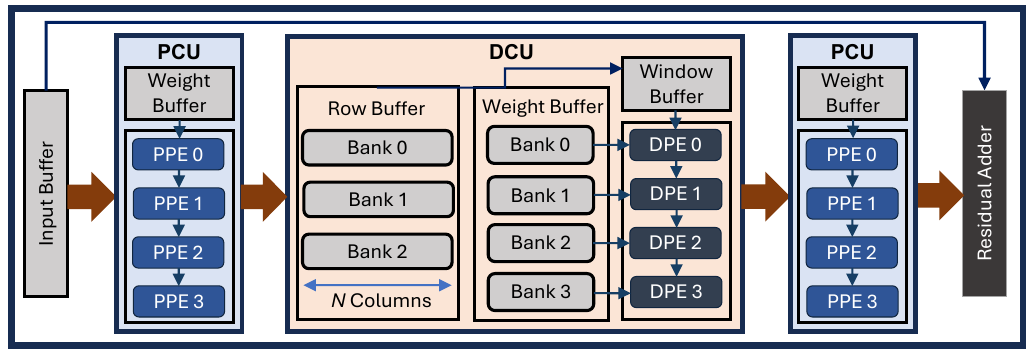}
    \caption{Streaming Architecture of Inverted Residual Block}
    \label{fig:streaming_mobilenet_block}
\end{figure}

\subsection{Pointwise Convolution Unit (PCU)}

For a given layer $l$ in the inverted residual block, the PCU is instantiated twice: first as an expansion layer that increases features from $\mathrm{InCh}_l$ to $\mathrm{ExpCh}_l$, and later as a projection layer that compresses features from $\mathrm{ExpCh}_l$ to $\mathrm{OutCh}_l$. The PCU comprises $p$ pixel-parallel processing elements (PPEs), each operating on one pixel and its associated channel vector. For each pixel, a PPE computes $m$ output channels in parallel, corresponding to $m$ distinct pointwise convolution filters. Each output channel is produced via a dot product between the input feature vector and its corresponding filter weights, with accumulation over input channels.

The weights are streamed from off-chip memory using contiguous transfers that match the FPGA platform’s native memory-port width, with multiple outstanding reads to keep the memory interface saturated despite variable off-chip DRAM latency. The streamed weights are stored on-chip and reused across computations within a tile, which is the basic unit of spatial computation within a layer, reducing off-chip memory traffic. To support parallel computation across output channels, weights are partitioned along the output-channel dimension using HLS array partitioning.

\subsection{Depthwise Convolution Unit (DCU)}

The DCU implements channel-wise $3 \times 3$ convolutions, where each input channel is convolved independently with its own $3 \times 3$ filter. The DCU operates on a streaming input and processes pixels sequentially. The DCU employs $n$ channel-parallel depthwise processing elements (DPEs). The input channel dimension is partitioned across DPEs, with each DPE responsible for a disjoint subset of $\mathrm{Ch}/n$ channels. All DPEs operate concurrently on the same $3 \times 3$ spatial window while computing depthwise convolutions for different channels using independent filters. 
Similar to PCU, depthwise convolution weights are streamed from off-chip memory and stored on-chip, then block-partitioned across the channel dimension using HLS array partitioning. 
A pixel at spatial location $(r, c)$ produces a valid output only if both row and column indices satisfy the stride condition, i.e., $r \bmod \text{stride} = 0$ and $c \bmod \text{stride} = 0$. The convolution for a pixel at spatial location $(r, c)$ requires access to pixels from rows $r-1, r,$ and $r+1$, and columns $c-1, c,$ and $c+1$. To provide this spatial context, the DCU employs a combination of row buffers and a sliding window buffer:

\paragraph{Row Buffers}
The DCU maintains three row buffers corresponding to rows $r-1, r,$ and $r+1$. Each row buffer stores an entire input row and is implemented using BRAM to provide sufficient storage capacity and efficient sequential access across the input width. As pixels stream in from the input, the row buffers advance vertically: once all columns of row $r+1$ have been consumed, the buffers shift such that the former $r$ row becomes $r-1$, the former $r+1$ row becomes $r$, and the newly streamed row is inserted as the new $r+1$.

\paragraph{Sliding Window Buffer}
To capture horizontal spatial neighbors, the DCU uses a sliding window buffer implemented with registers. For each row buffer, the sliding window tracks three adjacent column positions corresponding to columns $c-1, c,$ and $c+1$. As pixels stream across a row, the sliding window shifts by one column per cycle: previously stored center pixels become left neighbors, new pixels enter as right neighbors, and the window advances horizontally. Registers are used for the sliding window buffer to support cycle-level shifting and parallel, low-latency access to neighboring pixels across processing elements. Together, the row buffers and sliding window buffer assemble a complete $3 \times 3$ neighborhood for each pixel at location $(r, c)$.





\subsection{Hyperdimensional Encoding Unit (HEU)}
\label{subsec:hdc_encoding_engine}

\begin{figure}
    \centering
    \includegraphics[width=\columnwidth]{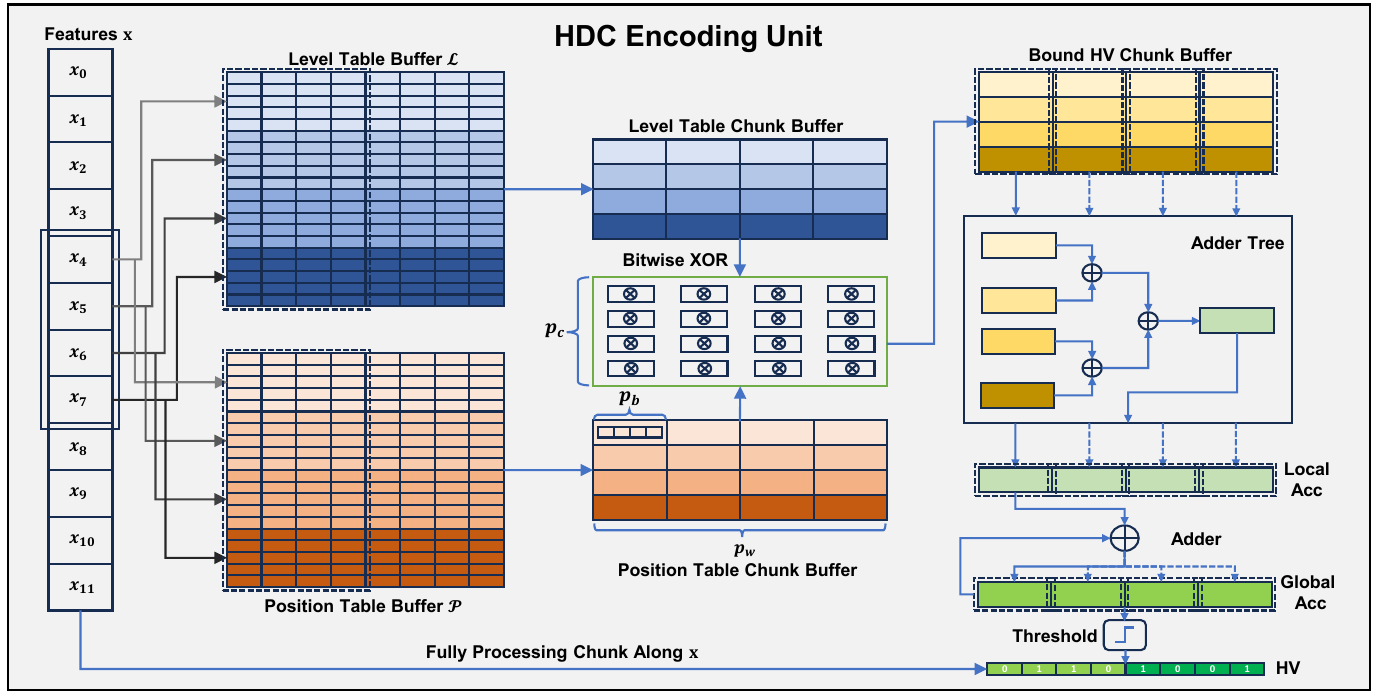}
    \caption{Hyperdimensional Encoding Unit (HEU)}
    \label{fig:hdc_encoding_engine}
\end{figure}

The HEU maps a quantized feature vector $\mathbf{x} \in \mathbb{R}^{F}$ into a binary hypervector (see Section~\ref{subsec:imagehd_encoding}).

\paragraph{Parallelization Strategy}
The HEU is organized as a 2D processing-element (PE) mesh of size $p_c \times (p_w \times p_b)$. Parallelism along the feature dimension ($F$) is exposed through $p_c$ independent feature lanes, while parallelism along the hypervector dimension ($D$) is achieved by packing hypervectors into words of $p_b$ bits and processing $p_w$ words concurrently.

\paragraph{On-Chip Item Memories and Banking}
Both the level table $\mathcal{L}$ and the position table $\mathcal{P}$ are stored on-chip in packed form. To enable conflict-free access across feature lanes, the position table is statically reorganized and banked across $p_c$ banks. This enables fully concurrent access to $\mathcal{P}$ for each feature lane. The level table is also stored on-chip and banked; since level indices are determined at runtime, a lightweight arbitration mechanism is used to handle concurrent accesses.

\paragraph{Binding and Local Accumulation}
In each cycle, the HEU processes a group of $p_c$ features and fetches the corresponding packed level and position hypervector words. Each feature lane performs element-wise binding using bitwise XOR, generating $p_c$ partial hypervectors of size $p_w \times p_b$ bits. These partial hypervectors are accumulated into per-chunk vote buffers using an adder tree that reduces bit-votes across the $p_c$ lanes at the word level. The reduced partial sums are merged into a global chunk accumulator, avoiding a single wide reduction across full $D$-dimensional hypervector.

\paragraph{Streaming Bundling and Thresholding}
The HEU operates in a chunk-wise streaming mode over the hypervector dimension. For each hypervector chunk of size $p_w \times p_b$, the HEU iterates across the feature dimension $F$ in groups of $p_c$ features. At each step, the corresponding level and position hypervector chunks are bound and accumulated into the chunk’s vote buffers. After all $\lceil F / p_c \rceil$ feature groups have contributed, the chunk is fully accumulated and a threshold operation is applied to perform majority voting. The resulting chunk is immediately written to its corresponding location in the global hypervector buffer and streamed to the HLU.

\subsection{Hyperdimensional Learning Unit (HLU)}
\label{subsec:classification_engine}

The HLU performs streaming similarity evaluation and online cluster updates on hypervector chunks produced by the HEU. It is organized as a $p_k \times (p_w \times p_b)$ PE mesh, enabling parallel processing across $p_k$ clusters per chunk. For each chunk, the HLU computes bitwise XORs between the encoded sample and stored cluster hypervectors, followed by population counts. Partial mismatch counts are accumulated into a global distance buffer of size $|\mathcal{C}|$, which yields full Hamming distances after all chunks are processed. A simple argmin selects the assigned cluster. Based on the admission decision, the HLU either initializes a new cluster or updates the matched cluster via chunk-wise bundling, while updating scalar cluster statistics online. The HLU is fully pipelined with the HEU, enabling streaming classification and cluster updates.

\subsection{Cluster Merge Unit (CMU)}
\label{subsec:merge_kmeanspp}

\begin{figure}
    \centering
    \includegraphics[width=\columnwidth]{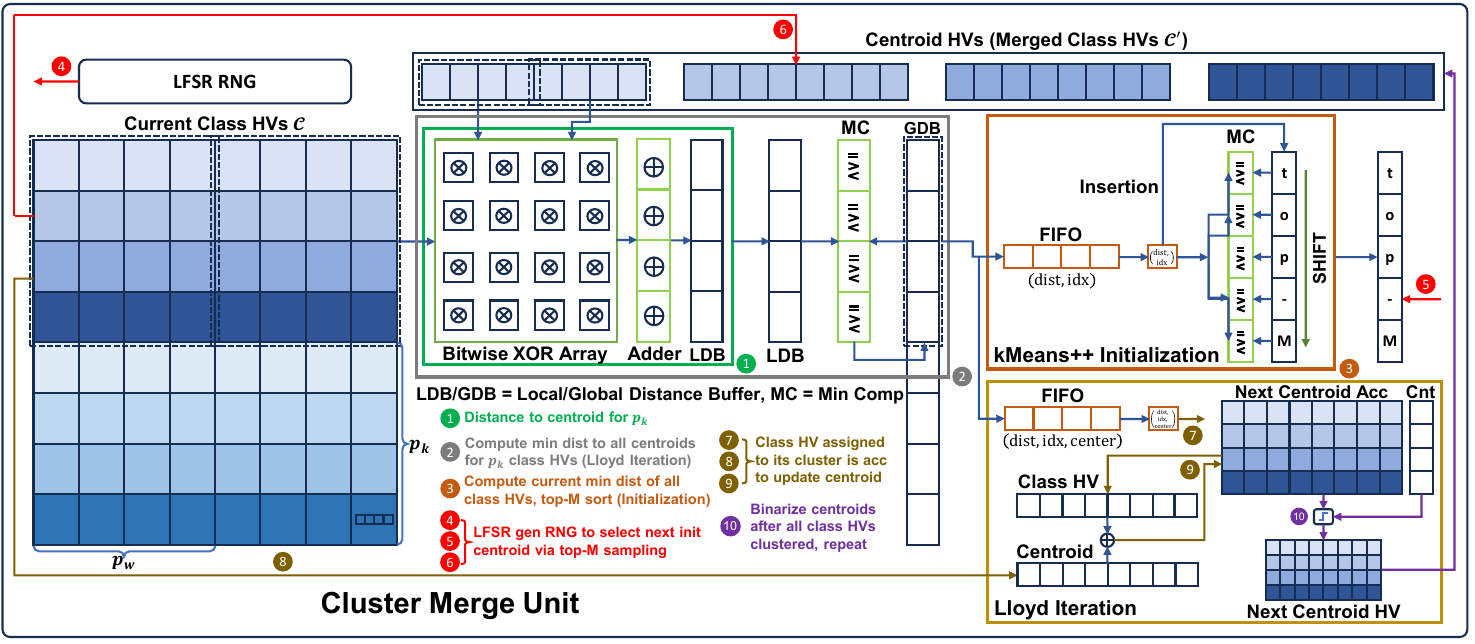}
    \caption{Cluster Merge Unit (CMU)}
    \label{fig:cluster_merge_engine}
\end{figure}

ImageHD invokes \textsc{MergeClusters} periodically (Algorithm~\ref{alg:imagehd_mergeclusters}) to consolidate the current prototype set.

\paragraph{Architecture and Dataflow}
The CMU is implemented as a 2D PE mesh of size $p_k \times (p_w\times p_b)$, similar to the HLU. The CMU streams class HVs chunk-wise over $D$ and performs XOR and popcount against an active centroid chunk, locally accumulating per-class partial Hamming distances and reducing them into full distances after $\lceil D/(p_w p_b)\rceil$ chunks are processed. 
\paragraph{Centroid Seeding and Top-$M$ Selection}
During Top-$M$ kMeans++ seeding, the CMU maintains a minimum-distance buffer $\mathbf{d}\in\mathbb{N}^{K}$ (stored in on-chip memory) that tracks each class's minimum Hamming distance to the currently selected centroid set. After selecting the first centroid, the CMU streams all $K$ class HVs through the above 2D mesh to compute distances and updates $\mathbf{d}$ via $p_k$ parallel comparators using a point-wise minimum. The Top-$M$ farthest candidates are identified using a compact streaming insertion buffer that maintains $M$ $(\text{dist},\text{idx})$ pairs, implemented with a comparator chain (of $M$ comparators) and a small shift network. Random sampling is driven by a lightweight 32-bit Linear Feedback Shift Register (LFSR), used for both initial centroid selection and uniform sampling within the Top-$M$ candidate set.

\paragraph{HV-Space Refinement and Centroid Update}
For refinement, assignments are computed by scanning current centroids sequentially and tracking the running argmin per class HV; assignment labels are queued and consumed by a bundling engine. Centroids are updated chunk-wise via HDC-style bundling and thresholding: per-bit votes are accumulated over class HVs assigned to their respective clusters, and binarized using a majority threshold derived from the assignment count. This design preserves binary centroid storage and avoids materializing full-precision accumulators over the $D$-bit HV.

\subsection{Compute Flow}

\begin{figure*}
    \centering
    \includegraphics[width=0.7\linewidth]{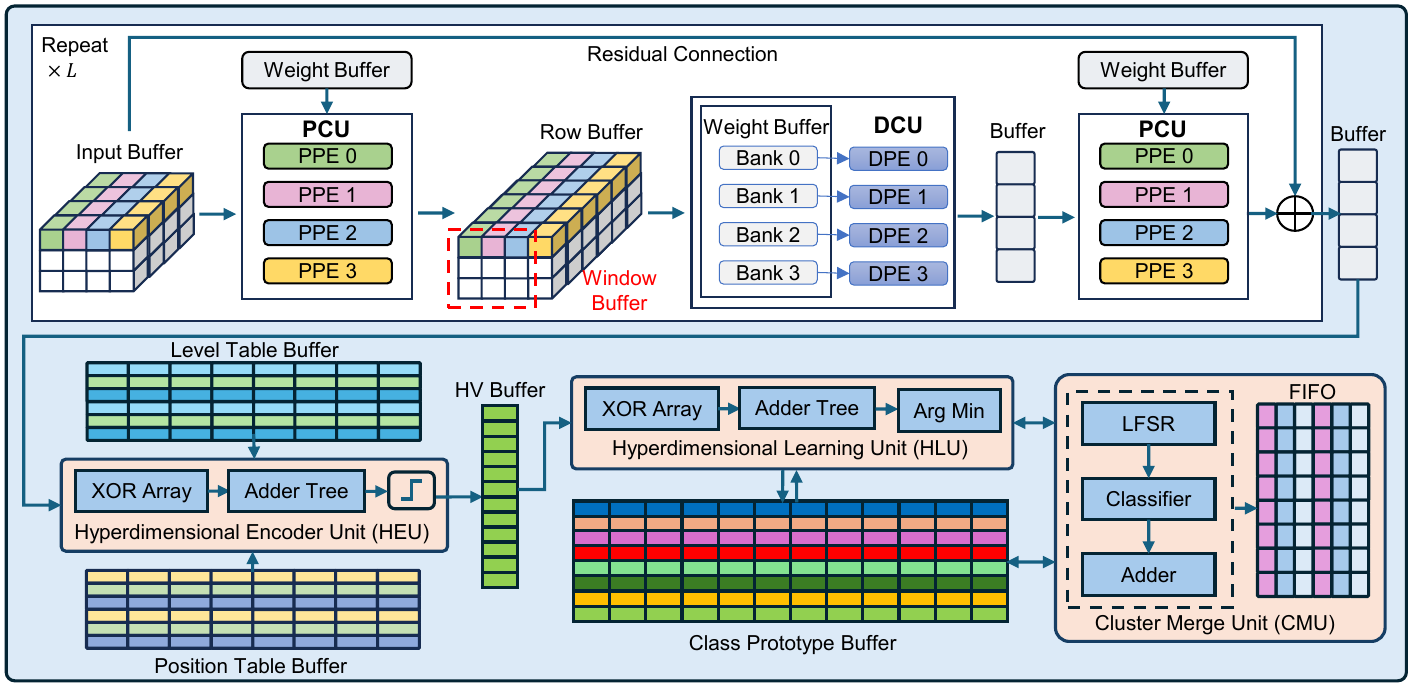}
    \caption{Compute Flow of our On-Device Continual Learning Accelerator}
    \label{fig:imagehd_compute_flow}
\end{figure*}

Figure~\ref{fig:imagehd_compute_flow} illustrates the end-to-end execution flow of ImageHD, integrating the specialized compute engines. The process begins with an input image of size $H \times W$, which is partitioned into spatial tiles of size $T_H \times T_W$ and processed once by a dedicated $3\times3$ convolution engine. The resulting feature maps are then tiled and streamed through a residual block consisting of the PCU and DCU, and this process is repeated for $L$ inverted residual layers. At layer $l$ and tile $(t_x, t_y)$, feature extraction is performed by an inverted residual block composed of expansion PCU, DCU, and projection PCU stages. 
If $\text{stride}=1$, the projection PCU output is combined with a buffered input activation stream to form a residual connection. All stages within the block operate in a streaming dataflow architecture and are connected via FIFOs, enabling pipelined execution. After all tiles of layer $l$ have been processed, the resulting feature map is stored on-chip and used as input to layer $l+1$, and the procedure repeats for $L$ layers. After this, a final $1\times1$ convolution is implemented using the PCU, and the output is streamed into the HEU to encode each feature vector $\mathbf{x}$ into a hypervector using level-position binding and majority bundling. Encoded hypervector chunks are streamed to the HLU, which computes Hamming distances against stored cluster prototypes and updates them. To bound memory growth, the CMU periodically consolidates prototypes using a hardware-efficient kMeans++.

\section{Experimental Evaluation}

\subsection{Implementation Details}
We implement the proposed accelerator on an AMD Zynq UltraScale+ ZCU104 FPGA~\cite{zcu104}, representative of an edge-class FPGA. 
The accelerator is designed using AMD Vitis High Level Synthesis (HLS) and synthesized using AMD Vitis Unified IDE v2024.2~\cite{AMD_Vitis_Unified_IDE_2024_2}. We instantiate the pointwise-depthwise convolution engine, which operates on fixed-size spatial tiles of size $\text{T}_H \times \text{T}_W$. In our implementation, $\text{T}_H = \text{T}_W = 32$. The engine exposes fixed parallelism along the pixel and channel dimensions: $p = 2$ pixels are processed in parallel, the pointwise convolution computes $m = 4$ output channels concurrently per pixel, and the depthwise convolution processes $n = 4$ channels in parallel. The HEU, HLU, and CMU are implemented with fixed parameters: $p_c=16$ feature lanes during encoding, $p_k=16$ cluster processing elements during similarity search, and hypervector chunks of $p_w=4$ words with $p_b=64$ bits per word. This configuration provides a balanced trade-off between latency, accumulator width, and on-chip resource utilization on ZCU104. 


\begin{table}
  \centering
  \caption{Resource Utilization of the Proposed Design on ZCU104}
  \label{tab:util}
  \resizebox{0.8\columnwidth}{!}{
  \begin{tabular}{lrrr}
    \toprule
    \textbf{Resource} & \textbf{Used} & \textbf{Available} & \textbf{Utilization} \\
    \midrule
    LUT          & 143{,}927 & 230{,}400 & 62\% \\
    FF           & 116{,}019 & 460{,}800 & 25\% \\
    BRAM (18K)   & 322       & 624       & 52\% \\
    DSP          & 126       & 1{,}728   & 7\%  \\
    URAM         & 57        & 96        & 59\% \\
    \bottomrule
  \end{tabular}}
\end{table}

\subsection{Benchmark Dataset}

\begin{table}
\centering
\caption{Dataset Statistics}
\label{tab:datasets}
\resizebox{0.8\columnwidth}{!}{
\begin{tabular}{lrrrr}
\toprule
\textbf{Dataset} & \textbf{\#Train} & \textbf{\#Test} & \textbf{Classes} & \textbf{Resolution} \\
\midrule
CIFAR-10   & 50{,}000  & 10{,}000 & 10                & $32 \times 32$ \\
CIFAR-100  & 50{,}000  & 10{,}000 & 20 (coarse)       & $32 \times 32$ \\
CORe50     & 120{,}000 & 45{,}000 & 50   & $128 \times 128$ \\
\bottomrule
\end{tabular}
}
\end{table}

We evaluate ImageHD on three benchmark datasets commonly used in prior continual learning works~\cite{9556356, ILOC, AILIS, LifeHD}: CIFAR-10, CIFAR-100~\cite{krizhevsky2009learning}, and CORe50~\cite{lomonaco2017core50}. The details of the datasets are summarized in Table~\ref{tab:datasets}, and we follow the train/test split suggested in~\cite{krizhevsky2009learning, lomonaco2017core50}.

\begin{table*}
\centering
\caption{Comparison with Prior FPGA-Based Continual Learning Accelerators}
\label{tab:prior_work}
\resizebox{0.8\linewidth}{!}{%
\begin{tabular}{lccccccc}
\toprule
\textbf{Prior Work} &
\textbf{Method} &
\textbf{Latency} &
\textbf{Power} &
\textbf{Energy Eff.} &
\textbf{Dataset} &
\textbf{Frequency} &
\textbf{Hardware} \\
 &  & \textbf{(ms)} & \textbf{(W)} & \textbf{(mJ/img)} &  & \textbf{(MHz)} &  \\
\midrule
\cite{9556356} 
& SLDA + ResNet-18
& 5.8837 & 8.69 & 51.1 & CORe50 & 200 & ZCU102 \\

AILIS \cite{AILIS}
& kNN + VGG-16
& 0.8163 & 6.035 & 4.92 & CORe50 & 200 & Zynq UltraScale+ MPSoC \\

ILOC \cite{ILOC}
& BAM-based Learning
& 3.010 & 22 & 66.22 & CIFAR-10 & 204 & Virtex-7 VU13P \\

\cite{LifeHD}
& Spectral-HDC + MobileNetV2
& 9.380 & -- & 93.75 & CIFAR-100 & -- & Jetson TX2 \\

\textbf{Ours}
& \textbf{kMeans++ HDC (End-to-End)}
& \textbf{0.4769} & \textbf{3.7} & \textbf{1.76} & \textbf{CORe50} & \textbf{300} & \textbf{ZCU104} \\
\bottomrule
\end{tabular}}
\end{table*}

\begin{figure}
    \centering
    \includegraphics[width=0.8\columnwidth]{Images/execution_time_comparison.pdf}
    \caption{%
    Execution time comparison of Spectral clustering (GPU) and \textbf{our} hardware-efficient kMeans++ consolidation algorithm on GPU and FPGA.
    }
    \label{fig:clustering_runtime}
\end{figure}
\subsection{Baseline Platforms}

\begin{table}
  \centering
  \caption{Specifications of Baseline Platforms}
  \label{tab:baseline-platforms}
  \resizebox{0.9\columnwidth}{!}{%
  \begin{tabular}{lccc}
    \toprule
    & \textbf{CPU} & \textbf{GPU} & \textbf{FPGA (Ours)} \\
    \midrule
    Platform & AMD EPYC 7763 & NVIDIA RTX A4000 & AMD ZCU104 \\
    Frequency & 2.45 GHz & 1560 MHz & 300 MHz \\
    Peak Perf. & 5.0 TFLOPS (FP32) & 19.2 TFLOPS (FP32) & 0.26 TFLOPS (FP32) \\
    On-chip Mem. & 32 MB L2, 256 MB L3 & 6 MB L2 Cache & 4.5 MB \\
    Memory BW & 409.6 GB/s (DDR4) & 448 GB/s (GDDR6) & 19.2 GB/s (DDR4) \\
    \bottomrule
  \end{tabular}}
\end{table}

We evaluate our FPGA accelerator against CPU and GPU baselines: the AMD EPYC 7763, a dual-socket processor with 128 cores, and the NVIDIA RTX A4000, a workstation-class GPU with 6144 CUDA cores. For deployment relevance, we additionally compare our FPGA accelerator against an edge-class GPU, Jetson AGX Orin. All baselines are implemented in PyTorch (v2.0.1) with Python~3.8 and CUDA~12. 

\subsection{Evaluation Metrics}
We report four key metrics:  
(i) \emph{Clustering performance}: evaluated using Accuracy (ACC), Purity, and Normalized Mutual Information (NMI), which measure optimal cluster-to-class matching, cluster homogeneity, and agreement between cluster assignments and true class labels, respectively~\cite{pmlr-v48-xieb16}; 
(ii) \emph{Latency} (ms/img): average per-image end-to-end training latency during continual learning at batch size 1, including feature extraction, HDC encoding, cluster prototype update, and cluster merging. We use the integrated analyzer in the AMD Vitis Unified IDE v2024.2 toolchain for FPGA measurements and PyTorch’s built-in timing library for CPU/GPU measurements;
(iii) \emph{Power} (W): average device power during steady-state continual learning at batch size 1, including static and dynamic components. We use the AMD Vitis Unified IDE tools, nvidia-smi, and powertop to obtain power measurements on the FPGA, GPU, and CPU, respectively;
(iv) \emph{Energy Efficiency} (mJ/img): average energy consumed per image during continual learning, computed as the product of measured power and per-image latency. 

\subsection{Effect of Quantization}
We evaluate post-training quantization using the ONNX library~\cite{onnx_quantization}. Averaged across datasets, INT8 quantization improves clustering accuracy and NMI over the FP32 by $+0.0120$ and $+0.0093$, while reducing purity by $-0.0535$. Compared to FP16, INT8 achieves comparable clustering accuracy ($-0.0049$) while improving purity and NMI by $+0.0206$ and $+0.0047$. Despite known limitations of post-training quantization for models such as MobileNetV2~\cite{8578384}, our results show that quantization does not degrade clustering performance. This confirms our hypothesis that the neural network's quantization noise is effectively tolerated by the downstream HDC encoder due to its robustness. We therefore select INT8 for hardware deployment, as it provides a favorable trade-off between clustering quality and a $3.7\times$ reduction in model size relative to FP32.

\begin{table}
\centering
\caption{Effect of Quantization on Clustering Performance}
\label{tab:quantization_results}
\resizebox{0.8\columnwidth}{!}{%
\begin{tabular}{llcccc}
\toprule
\textbf{Dataset} & \textbf{Precision} & \textbf{ACC} & \textbf{Purity} & \textbf{NMI} & \textbf{Model Size (MB)} \\
\midrule
\multirow{3}{*}{CIFAR-10}
 & FP32 & 0.2528 & \textbf{0.3660} & 0.1138 & 8.5 \\
 & FP16 & \textbf{0.2792} & 0.1746 & 0.1153 & 4.3 \\
 & INT8 & 0.2689 & 0.3231 & \textbf{0.1327} & 2.3 \\
\midrule
\multirow{3}{*}{CIFAR-100}
 & FP32 & 0.1846 & \textbf{0.1769} & 0.1248 & 8.5 \\
 & FP16 & 0.1902 & 0.1674 & 0.1252 & 4.3 \\
 & INT8 & \textbf{0.1940} & 0.1619 & \textbf{0.1283} & 2.3 \\
\midrule
\multirow{3}{*}{CORe50}
 & FP32 & 0.1839 & \textbf{0.2732} & 0.0563 & 8.5 \\
 & FP16 & \textbf{0.2026} & 0.2518 & \textbf{0.0681} & 4.3 \\
 & INT8 & 0.1944 & 0.1707 & 0.0618 & 2.3 \\
\bottomrule
\end{tabular}}
\end{table}

\subsection{Effect of Unified Exemplar Memory}

We compare the memory footprint of a two-tier cluster memory with working memory and long-term memory used in~\cite{LifeHD,Baddeley1992WorkingMemory} against our unified exemplar design. Both designs are implemented in PyTorch to evaluate memory usage. The unified design reduces the worst-case memory footprint for cluster storage from 189.3 KB to 126.6 KB, a 33\% reduction.

\subsection{Effect of Our Prototype Consolidation Algorithm}

We evaluate the performance of our hardware-efficient kMeans++ consolidation algorithm by measuring the total clustering execution time across the entire continual training stream for each dataset. Table~\ref{tab:clustering_results} and Figure~\ref{fig:clustering_runtime} summarize these results, capturing the impact of algorithmic and hardware optimizations. Comparing spectral clustering and k-means on GPU highlights the effect of algorithmic optimization: spectral clustering incurs high computational cost due to eigenvalue decomposition with cubic complexity, making it unsuitable for continual learning. Therefore, compared to spectral clustering, k-means yields an average speedup of \textbf{8-9$\times$} across datasets, with only negligible average reductions in clustering accuracy ($-0.0035$), purity ($-0.0468$), and NMI ($-0.0006$). We further compare k-means execution on GPU and FPGA to highlight the effect of architectural optimizations. Despite the GPU's higher compute power, the FPGA achieves lower latency through our design, which combines simplified k-means initialization via Top-$M$ uniform sampling, a small number of Lloyd refinement iterations, and a streaming dataflow mapped onto a two-dimensional PE array. As a result, FPGA achieves up to \textbf{145.7$\times$} speedup over spectral clustering.

\subsection{End-to-End Performance Comparison}

We compare our FPGA accelerator against CPU and GPU baselines with the same algorithmic optimizations discussed in Section~\ref{section-4}, including an identical clustering strategy and INT8 quantization for the feature extractor. All platforms execute under a batch-1 configuration to reflect the online continual learning setting. While the CPU and GPU baselines offer significantly higher peak compute capability and memory bandwidth, our accelerator achieves up to \textbf{40.4$\times$} and \textbf{4.84$\times$} lower latency compared to the CPU and GPU baselines, respectively, on the CORe50 dataset. We further evaluate an embedded GPU baseline using a Jetson AGX Orin Developer Kit to represent edge deployment scenarios. 
The Jetson platform exhibits significantly higher latency and energy per sample compared to our FPGA design under identical algorithmic conditions.

These improvements are due to the streaming dataflow architecture, which overlaps computation and data fetch in the compute-intensive inverted residual blocks, as well as streaming, word-packed hardware implementations of hyperdimensional encoding, similarity search, and bounded online cluster consolidation that exploit parallel bitwise operations. 

\begin{table}
\centering
\caption{End-to-End Performance and Energy Comparison}
\label{tab:perf_energy_all}
\scriptsize
\setlength{\tabcolsep}{4pt}
\renewcommand{\arraystretch}{1.05}
\resizebox{0.8\columnwidth}{!}{%
\begin{tabular}{llccccc}
\toprule
\textbf{Dataset} &
\textbf{Device} &
\textbf{Latency} &
\textbf{Throughput} &
\textbf{Power} &
\textbf{Energy Eff.} \\
 &  & \textbf{(ms)} & \textbf{(img/s)} & \textbf{(W)} & \textbf{(mJ/img)} \\
\midrule
\multirow{3}{*}{CIFAR-10}
& CPU  & 7.60 (1$\times$) & 131.58 & 33 & 250.80 (1$\times$) \\
& GPU  & 0.96 (7$\times$) & 1041.67 & 78 & 74.88 (3$\times$) \\
& Jetson & 12.58 (0.6$\times$) & 79.49 & 5.6 & 70.4 (3.6$\times$) \\

& \textbf{FPGA (Ours)} & \textbf{0.27 (28$\times$)} & \textbf{3703.70} & \textbf{3.6} & \textbf{0.97 (258$\times$)} \\
\midrule
\multirow{3}{*}{CIFAR-100}
& CPU  & 9.10 (1$\times$) & 109.89 & 35 & 318.50 (1$\times$) \\
& GPU  & 1.02 (8$\times$) & 980.39 & 77 & 78.54 (4$\times$) \\
& Jetson & 16.96 (0.5$\times$) & 58.96 & 5.29 & 89.7 (3.5$\times$) \\
& \textbf{FPGA (Ours)} & \textbf{0.34 (26$\times$)} & \textbf{2965.77} & \textbf{3.6} & \textbf{1.21 (263$\times$)} \\
\midrule
\multirow{3}{*}{CORe50}
& CPU  & 19.25 (1$\times$) & 51.95  & 35 & 673.75 (1$\times$) \\
& GPU  & 2.31  (8$\times$) & 432.90 & 80 & 184.80 (3$\times$) \\
& Jetson & 13.14 (1.5$\times$) & 76.10 & 5.9 & 77.5 (8.7$\times$) \\
& \textbf{FPGA (Ours)} & \textbf{0.48 (40$\times$)} & \textbf{2097.05} & \textbf{3.7} & \textbf{1.76 (383$\times$)} \\
\bottomrule
\end{tabular}}
\end{table}

\begin{table}
\centering
\caption{Effect of our Clustering Algorithm on Accuracy}
\label{tab:clustering_results}
\resizebox{0.8\columnwidth}{!}{%
\begin{tabular}{llccccc}
\toprule
\textbf{Dataset} & \textbf{Method} & \textbf{Accuracy} & \textbf{Purity} & \textbf{NMI} \\
\midrule
CIFAR-100 & Spectral        & \textbf{0.1846} & 0.1769 & \textbf{0.1248}\\
 & \textbf{k-means (Ours)}         & 0.1703 & \textbf{0.2006} & 0.1193 \\
\midrule
CIFAR-10  & Spectral        & \textbf{0.2528} & \textbf{0.3660} & 0.1138\\
  & \textbf{k-means (Ours)}         & 0.2524 & 0.2281 & \textbf{0.1149}\\
\midrule
CORe50    & Spectral        & 0.1839 & \textbf{0.2732} & 0.0563\\
    & \textbf{k-means (Ours)}         & \textbf{0.1881} & 0.2471 & \textbf{0.0589}\\
\bottomrule
\end{tabular}}
\end{table}

\subsection{Comparison with Prior FPGA Accelerators}
\label{prior_fpga_acc}
We compare ImageHD against representative FPGA and edge accelerators for continual and incremental learning in supervised settings, as there are no prior FPGA accelerators targeting unsupervised continual learning. ImageHD achieves \textbf{19.7$\times$} lower end-to-end latency than~\cite{LifeHD}, which operates on an edge GPU using a MobileNetV2 backbone and spectral clustering. In addition, ImageHD improves energy efficiency by \textbf{53.27$\times$} compared to~\cite{LifeHD}. These gains result from our hardware-efficient kMeans++ consolidation strategy and the streaming dataflow architecture. Compared to ILOC~\cite{ILOC}, ImageHD achieves \textbf{6.3$\times$} lower latency and \textbf{37$\times$} higher energy efficiency. ILOC reports latency only for the associative memory-based learning stage and excludes feature extraction, which constitutes a major portion of the end-to-end latency. In contrast, ImageHD reports end-to-end latency.~\cite{9556356} and AILIS~\cite{AILIS} target supervised continual learning and employ heavier backbone models (ResNet-18 and VGG-16). AILIS further reports feature-extraction latency from prior work evaluated on CIFAR-10, a smaller dataset than CORe50. Despite operating larger dataset, ImageHD achieves lower latency. Direct accuracy comparisons to these accelerators are not applicable, since they focus on supervised learning. 
\section{Related Works}

\textbf{Continual Learning for Visual Data:} 
Continual learning addresses the challenge of incrementally learning from sequential data streams without catastrophic forgetting~\cite{McCloskey1989Catastrophic, Parisi2019CLReview}. Early approaches focused on regularization-based methods such as Elastic Weight Consolidation (EWC)~\cite{Kirkpatrick2017EWC} and Synaptic Intelligence~\cite{Zenke2017SI}, as well as replay-based methods such as iCaRL~\cite{Rebuffi2017iCaRL} and GEM~\cite{LopezPaz2017GEM}. Recent surveys~\cite{DeLange2022CLSurvey, Wang2023CLSurvey} analyze continual learning methods, highlighting the trade-off between plasticity and stability. Recent works focus on-device continual learning with bounded memory and without full retraining~\cite{OnDeviceTL}. Self-supervised methods such as CaSSLe~\cite{Fini2022CaSSLe}, LUMP~\cite{Madaan2022LUMP}, and Lifelearner~\cite{LifeLongLearner} reduce label requirements but rely on backpropagation. In contrast, ImageHD targets non-iterative on-device CL.


\textbf{Continual Learning Accelerators:} Hardware acceleration for continual learning remains largely unexplored. Deep SLDA~\cite{Hayes2020DeepSLDA} enables streaming class-incremental learning via linear discriminant analysis but lacks dedicated hardware support. Prior work accelerates exemplar-based or SLDA-style incremental learning on an FPGA but often excludes feature extraction or relies on heavyweight CNN backbones~\cite{AILIS, ILOC, 9556356}. Recent work on CNN FPGA acceleration~\cite{Guo2017FPGASurvey, Zhang2018Caffeine} provides foundations for efficient neural feature extraction on edge devices. ImageHD advances this area by providing the first end-to-end FPGA accelerator for \emph{unsupervised} continual learning, integrating INT8-quantized MobileNetV2, streaming HDC encoding, and bounded-memory cluster management.

\section{Conclusion}
We presented ImageHD, the first end-to-end FPGA accelerator for unsupervised continual image learning that tightly integrates a quantized neural feature extractor with a streaming hyperdimensional computing pipeline. By co-designing the algorithm and hardware, ImageHD enables fully on-device learning with bounded memory growth and low-latency inference. Experimental results demonstrate that ImageHD achieves strong accuracy while significantly improving latency and energy efficiency compared to CPU and GPU baselines.

\section{Acknowledgement}
This work is supported by the DEVCOM Army Research Lab (ARL) under grant W911NF-242-0194, and the National Science Foundation (NSF) under grants CSSI-2311870 and OAC-2505107.
The authors thank Haoyang Fan for assistance in setting up the Jetson AGX Orin platform. 



\bibliographystyle{IEEEtran}
\bibliography{Reference}

\end{document}